\documentclass[letterpaper]{article} 
\usepackage{aaai2026}  
\usepackage{times}  
\usepackage{helvet}  
\usepackage{courier}  
\usepackage[hyphens]{url}  
\usepackage{graphicx} 
\usepackage{natbib}  
\usepackage{caption} 
\usepackage{algorithm}
\usepackage{algorithmic}

\usepackage{newfloat}
\usepackage{listings}
\DeclareCaptionStyle{ruled}{labelfont=normalfont,labelsep=colon,strut=off} 
\floatstyle{ruled}
\newfloat{listing}{tb}{lst}{}
\floatname{listing}{Listing}
\usepackage{amsmath,amsfonts}
\usepackage{array}
\usepackage{multirow}
\usepackage{xcolor}
\title{VPHO: Joint Visual-Physical Cue Learning and Aggregation for Hand-Object Pose Estimation}
\author{
    Jun Zhou\textsuperscript{\rm 1,2,3,4},
    Chi Xu\textsuperscript{\rm 1,2,3}\thanks{Corresponding author},
    Kaifeng Tang\textsuperscript{\rm 1,2,3},
    Yuting Ge\textsuperscript{\rm 1,2,3},
    Tingrui Guo\textsuperscript{\rm 1,2,3},
    Li Cheng\textsuperscript{\rm 4}
}
\affiliations{
    \textsuperscript{\rm 1}School of Automation, China University of Geosciences, Wuhan 430074, China,\\
    \textsuperscript{\rm 2}Hubei Key Laboratory of Advanced Control and Intelligent Automation for Complex Systems, Wuhan 430074, China,\\
    \textsuperscript{\rm 3}Engineering Research Center of Intelligent Technology for Geo-Exploration, Ministry of Education, Wuhan 430074, China,\\
    \textsuperscript{\rm 4}Department of Electrical and Computer Engineering, University of Alberta, Edmonton, AB T6G 2R3, Canada\\
    \{jchow, xuchi, tkf, gyting, guotingrui\}@cug.edu.cn, lcheng5@ualberta.ca
}

\usepackage{bibentry}

\begin{document}

\maketitle

\begin{abstract}
    Estimating the 3D poses of hands and objects from a single RGB image is a fundamental yet challenging problem, with broad applications in augmented reality and human-computer interaction. 
    Existing methods largely rely on visual cues alone, often producing results that violate physical constraints such as interpenetration or non-contact. 
    Recent efforts to incorporate physics reasoning typically depend on post-optimization or non-differentiable physics engines, which compromise visual consistency and end-to-end trainability.
    To overcome these limitations, we propose a novel framework that jointly integrates visual and physical cues for hand-object pose estimation.
    This integration is achieved through two key ideas:
    1) joint visual-physical cue learning: The model is trained to extract 2D visual cues and 3D physical cues, thereby enabling more comprehensive representation learning for hand-object interactions;
    2) candidate pose aggregation: A novel refinement process that aggregates multiple diffusion-generated candidate poses by leveraging both visual and physical predictions, yielding a final estimate that is visually consistent and physically plausible.
    Extensive experiments demonstrate that our method significantly outperforms existing state-of-the-art approaches in both pose accuracy and physical plausibility.
\end{abstract}

\begin{links}
    \link{Code}{https://github.com/zhoujun-7/VPHO}
\end{links}

\section{Introduction}

\begin{figure}[t]
    \centering
    \includegraphics[width=0.9\columnwidth]{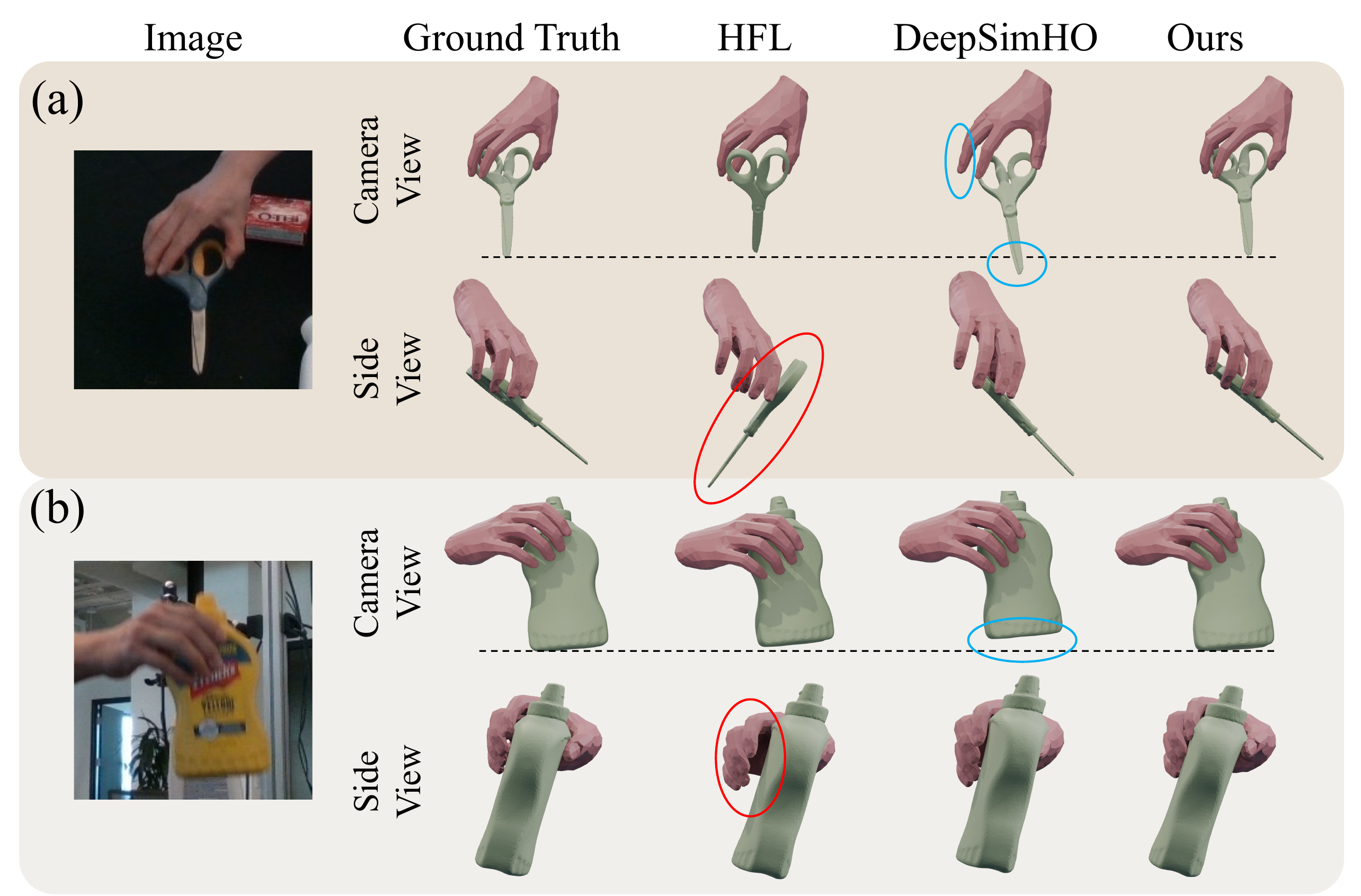}
    \caption {Visual comparison between a state-of-the-art visual-only method, HFL, and a method that incorporates physical cues, DeepSimHO. HFL yields visually aligned yet physically implausible results (red circles), while DeepSimHO improves physical plausibility at the cost of visual alignment (blue circles).}
    \label{fig_intro}
\end{figure}

Hand-object pose estimation from single RGB images~\cite{CVPR19_HOPE} has broad applications across various fields, including virtual and augmented reality~\cite{ToG19_VR, JPCS19_AR}, human-computer interaction~\cite{IJITDM20_CHI}, and robotics~\cite{Science19_Robotics}.
Most existing methods~\cite{CVPR22_KYPT, CVPR21_Semi, CVPR23_HFL, WACV23_DMA} primarily rely on image-based visual cues to ensure that the 3D pose estimation is consistent with 2D observations~\cite{CVPR24_SimpleBaseline, CVPR24_HAMER}.
For example, segmentation consistency losses are widely adopted~\cite{CVPR24_HOISDF, CVPR24_MOHO, CVPR23_H2Onet} to align projected 3D meshes with 2D segmentation masks, while photometric consistency across frames is explored in~\cite{CVPR20_Hasson}.
However, these visual-based methods often neglect physical constraints, leading to physically implausible artifacts such as penetration or lack of contact. As illustrated in Figure~\ref{fig_intro}, the result from HFL~\cite{CVPR23_HFL} appears visually reasonable in the original camera view but reveals improper grasping when rendered from a different viewpoint, as highlighted in the red circle.

To improve physical plausibility, several works~\cite{CVPR20_ContactPose, ECCV22_AlignSDF, TPAMI24_CPF} introduce post-optimization strategies that incorporate physical constraints.
While effective at reducing artifacts such as interpenetration or missing contacts, these methods are highly sensitive to the quality of the initial pose and often produce unstable or suboptimal outputs when the initialization is inaccurate.
To better balance visual and physical fidelity, more recent approaches~\cite{CVPR19_HOPE, CVPR20_LUKE, CVPR24_LEMON, CVPR23_H2OTR, NIPS23_DeepSimHO, SigGraph24_HOIC} propose end-to-end training methods that incorporate physical cues.
While this integration improves physical plausibility, it often comes at the cost of degraded visual consistency.
As shown in Figure~\ref{fig_intro}, DeepSimHO~\cite{NIPS23_DeepSimHO} improves physical realism over its visual-only baseline~\cite{CVPR22_ArtiBoost}; yet, the predicted object pose deviates from the image observation (blue circle), compromising pose accuracy.

To resolve these limitations, we propose a novel approach that effectively integrates visual and physical cues to ensure both visual consistency and physical plausibility.
This is accomplished through two key ideas:
1) Joint visual-physical cue learning: Our model is trained to extract 2D visual cues (e.g., hand and object heatmaps) alongside 3D physical cues (e.g., force vectors), enabling richer representation learning for hand-object interactions.
2) Candidate pose aggregation: We propose a novel aggregation process that leverages both predicted visual and physical cues to aggregate multiple candidate poses generated by a diffusion model into a single, physically plausible and visually consistent estimate.

A core challenge lies in the prediction of 3D interaction forces due to: (1) their high dimensionality, (2) the complexity of contact dynamics and friction modeling, and (3) the lack of ground-truth force annotations.
To address this, we introduce a \textbf{Force Prediction Module} that models local contact forces using friction cones and transforms them from local to global coordinates to compute the overall hand-object interaction force.
The module is trained via a semi-supervised strategy using physical constraints and pseudo force labels, without requiring ground-truth annotations.

Pose aggregation also poses unique challenges due to the high degrees of freedom of the pose parameters (e.g., articulated hand joints) and the interdependence of hand-object interaction.
To address this, we propose a two-stage aggregation scheme:
1) \textbf{Visual-based Aggregation}: Candidates are hierarchically aggregated along the kinematic chain using visual cues, which effectively reduces error accumulation and enhances visual consistency.
2) \textbf{Physics-based Aggregation}: Candidates are ranked and selected based on physical constraints, such as contact and torque balance, which improves contact quality and enhances physical plausibility.

\vspace{0.3em}

In summary, the contributions of this work are as follows:
\begin{itemize}
  \item We propose a novel hand-object pose estimation approach that integrates both visual and physical cues without compromising either.
  \item We introduce a force prediction module that models interaction forces via friction cones and local-to-global transformation, trained with physical constraints and pseudo force labels.
  \item We propose a two-stage pose aggregation strategy that leverages both visual and physical cues to yield accurate and physically plausible hand-object poses.
\end{itemize}
Extensive experiments on standard benchmarks demonstrate that our method achieves state-of-the-art performance in both pose accuracy and physical plausibility.

\section{Related Works}
Most existing efforts~\cite{CVPR23_H2Onet, CVPR20_Hasson, CVPR22_Collaborative, CVPR24_SimpleBaseline, CVPR25_WiLoR} estimate hand and object poses primarily by leveraging image-based visual cues, ensuring alignment between the 2D projections of the estimated 3D poses and the corresponding image observations.
Several methods~\cite{CVPR24_HOISDF, CVPR24_MOHO, CVPR23_H2Onet} incorporate segmentation losses to enforce consistency between the projected 3D mesh and 2D image segmentations.
Similarly, Park et al.~\cite{CVPR22_HandOccNet} employ a transformer module to inject hand information into occluded 2D regions.
%
%
However, these methods predominantly focus on visual information and do not explicitly incorporate physical constraints, making them susceptible to physically implausible predictions such as interpenetration or lack of contact.

To address this limitation, several methods~\cite{CVPR21_ContactOpt, TPAMI24_CPF, arXiv24_EgoPressure} propose post-optimization strategies that refine initial pose estimates by incorporating physical cues.
For instance, Grady et al.\cite{CVPR21_ContactOpt} and Tse et al.\cite{ECCV22_S2Contact} infer desirable hand-object contact regions from initial poses and subsequently optimize both the hand and object to better conform to these inferred regions.
%
%
Similarly, Hu et al.\cite{SigGraph22_Physical} refine hand-object interactions by adjusting fingertip forces and contact points based on initial motion trajectories.
While these methods are effective in improving physical plausibility, they rely heavily on accurate initializations. Inaccurate initial poses can cause divergence from visual evidence, resulting in visually inconsistent estimations.

\begin{figure*}[t]
    \centering
    \includegraphics[width=0.8\textwidth]{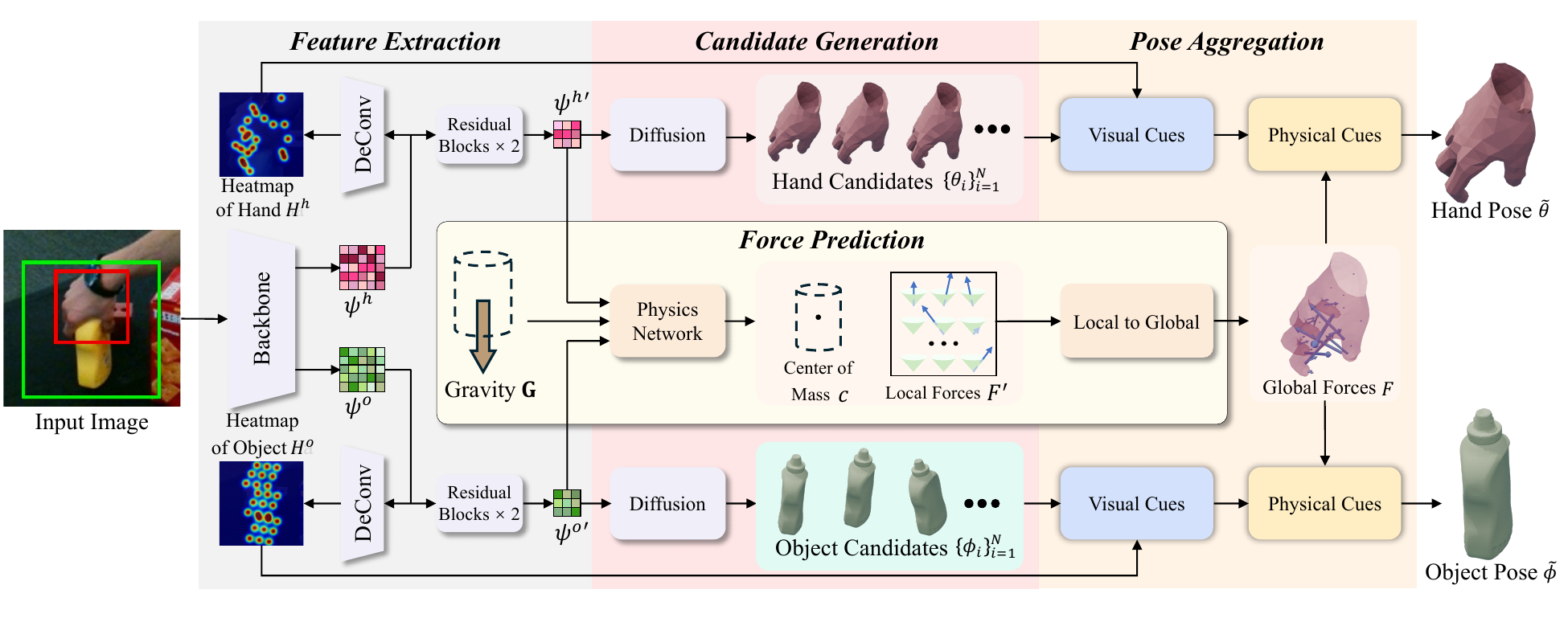} 
    \caption{
        The framework of our approach, consisting of the following four modules: feature extraction, force prediction, candidate generation and pose aggregation.}
    \label{fig_method_framework}
\end{figure*}

More recently, several studies have explored the unification of visual and physical reasoning within an end-to-end learning framework~\cite{CVPR19_HOPE, CVPR23_H2OTR, NIPS23_DeepSimHO, SigGraph24_HOIC}.
%
%
Hu et al.~\cite{AAAI24_CHOI} model part-level and vertex-level contact probabilities to construct an implicit neural representation of the object, thereby facilitating object pose inference.
Wang et al.~\cite{NIPS23_DeepSimHO} incorporate a physics engine into the training loop to supervise the learning of stability-aware poses based on simulated physical outcomes.
While these methods represent progress toward unified visual-physical modeling, they often compromise visual fidelity in favor of physical plausibility.
In contrast, our approach integrates visual and physical cues during inference by aggregating multiple candidate poses generated by a diffusion model, enabling the selection of solutions that are both visually consistent and physically plausible.

\textcolor{black}{
    Following~\cite{CVPR23_HFL,CVPR24_HOISDF}, we focus on estimating both hand pose and object 6D pose from a single RGB image, with camera intrinsics and the object CAD model being available. Methods such as \cite{ECCV24_WILD_HOI}, which reconstruct only the mesh of the manipulated object without estimating either the hand pose or object 6D pose, are therefore not considered in our comparison.
}

\section{Method}
The framework of our approach is illustrated in Figure~\ref{fig_method_framework}.

\subsection{Feature Extraction}
\label{sec_method_feature_extraction}
Given an input RGB image, an enhanced ResNet50 backbone network~\cite{CVPR23_HFL} is employed to extract features for both the hand and the object. 
These features are subsequently processed by DeConv layers~\cite{ECCV18_SimpleBaseline} to generate the hand heatmap $H^{h}$ and the object heatmap $H^{o}$, which serve as visual cues in pose aggregation module. 
Two residual blocks~\cite{CVPR16_ResNet} further refine the hand and object features, preparing them for force prediction and candidate pose generation. 
The loss function used to supervise the heatmap predictions is defined as:
\begin{equation}
    \label{equ_heatmap_loss}
      \begin{aligned}
      \mathcal{L}_{hm}&=\lambda_{hm}\left(\mathcal{L}_{mse}(H^{h}, \bar{H}^{h})+\mathcal{L}_{mse}(H^{o}, \bar{H}^{o})\right),
  \end{aligned}
  \end{equation}
where $\mathcal{L}_{mse}$ denotes the mean square error loss function, $H^{h} \in \mathbb{R}^{\vert J^{h}\vert\times h_{m}\times w_{m}}$ and $\bar{H}^{h}\in\mathbb{R}^{\vert J^{h}\vert\times h_{m}\times w_{m}}$ represent the predicted and ground truth heatmaps for the hand joints $J^{h}$, respectively. 
$H^{o} \in \mathbb{R}^{\vert J^{o}\vert\times h_{m}\times w_{m}}$ and $\bar{H}^{o}\in\mathbb{R}^{\vert J^{o}\vert\times h_{m}\times w_{m}}$ represent the predicted and ground truth heatmaps for the object keypoints $J^{o}$, respectively. 
$h_{m}$ and $w_{m}$ denote the height and width of the heatmap, and $\lambda_{hm}$ is a hyperparameter that weights the heatmap loss. Detailed settings of all hyperparameters are provided in the extended version.

\subsection{Force Prediction}
\label{sec_method_force_prediction}
\textcolor{black}{
    We focus on hand-object pose estimation from a single image, where motion-related quantities such as acceleration cannot be inferred. 
    Thus, we adopt the static equilibrium assumption as a necessary simplification, which is shown to be effective in our ablation studies.
    %
}

\begin{figure}[t]
    \centering
    \includegraphics[width=0.9\columnwidth]{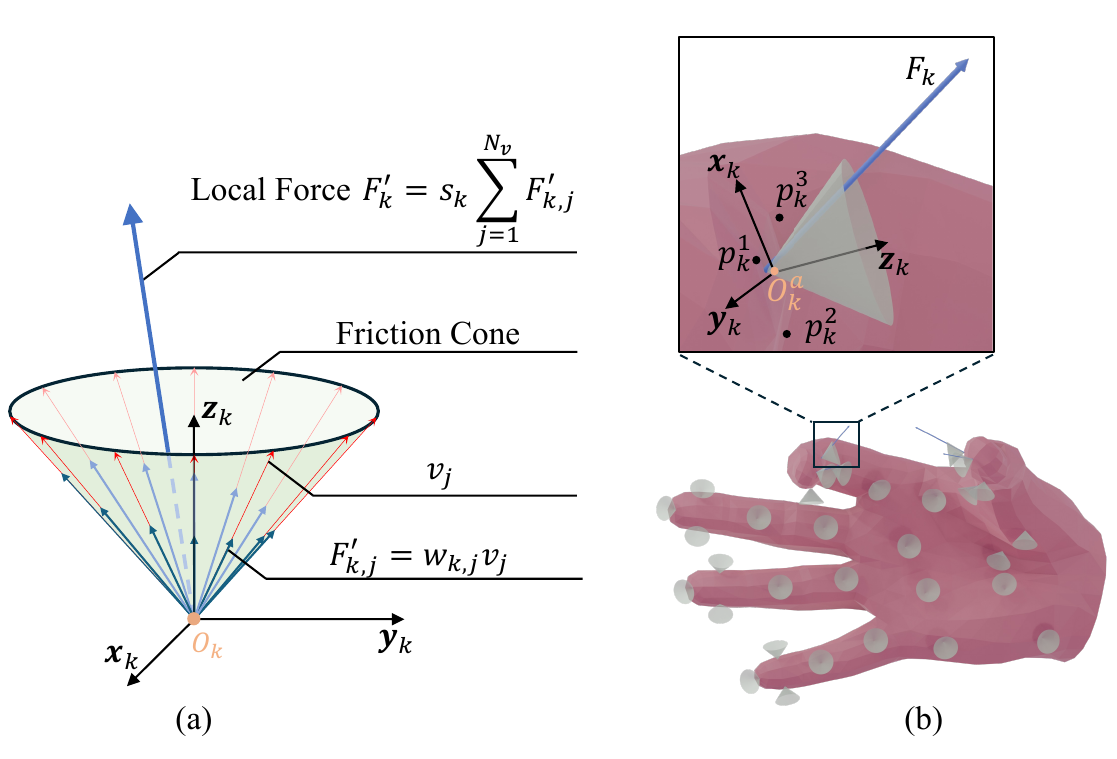}
    \caption {Friction cone and force representations in (a) local and (b) global coordinate frames.}
    \label{fig_method_local_global_force}
\end{figure}

\subsubsection{Local Force}
\label{sec_method_local_force}
Hand-object contact interactions are inherently complex~\cite{SigGraph22_Physical}. Following~\cite{TPAMI24_CPF}, we approximate this complexity by representing contact forces through 32 sparse anchor points$\left\{ O_{k}^{a}\right\}_{k=1}^{32}$ on the hand surface. Each anchor point $O_k^a$ is associated with a local coordinate system, as depicted in Figure~\ref{fig_method_local_global_force}(a).
According to Coulomb's friction law~\cite{CUP08_ClassicalMechanics}, contact forces must lie within a friction cone determined by a friction coefficient $\mu$. 
We model this cone using a set of base vectors $\left\{ v_{j}\right\} _{j=1}^{N_{v}}$, where each vector is defined as:
\begin{equation}
\label{equ_local_force1}
    v_{j}=\left(\mu\sin(\frac{2\pi j}{N_{v}}),\mu\cos(\frac{2\pi j}{N_{v}}),1\right). 
\end{equation}
The local force $F_{k}^{\prime}$ at anchor point $O_{k}^{a}$ is expressed as a weighted sum of these base vectors:
\begin{equation}
  \label{equ_local_force}
  F_{k}^{\prime}=s_{k}\sum_{j=1}^{N_{v}}w_{k,j}v_{j},
\end{equation}
where $s_k \in \mathbb{R}$ and $w_{k,j} \in \mathbb{R}$ are the learned scaling and weighting coefficients, respectively.

\subsubsection{Global Force}
\label{sec_method_global_force}
As illustrated in Figure~\ref{fig_method_local_global_force}(b), the local force $F_k^{\prime}$ is transformed into a global force $F_k$ using the MANO hand model~\cite{TOG17_MANO} through its linear blend skinning procedure.
Each anchor point $\left\{ O_{k}^{a}\right\} $ is attached to a triangle $\left\{ \triangle p_{k}^{1}p_{k}^{2}p_{k}^{3}\right\} $ on the hand mesh vertices $V^{h}$. The transformation is defined as:
\begin{equation}
  \label{equ_global_force}
  \left\{ \begin{aligned}
    F_{k}     &=\mathbf{R}_{k}^{L2G}F_{k}^{\prime}, \\
	O_{k}^{a} &=\sum_{i\in\{1,2,3\}}\alpha_{k}^{i}p_{k}^{i}.
	\end{aligned}\right.
\end{equation}
Here $F_k^{\prime}$ and $F_k$ represent the local and global force for the anchor point $O_k^a$, respectively.
$\mathbf{R}_{k}^{L2G}=[\mathbf{x}_{k},\mathbf{y}_{k},\mathbf{z}_{k}]$ denotes the rotation matrix, $\mathbf{x}_{k}$ is calulated by normalizing the vector $p_{k}^{2}-p_{k}^{1}$, $\mathbf{z}_{k}$ is calulated by normalizing the vector $(p_{k}^{2}-p_{k}^{1})\times(p_{k}^{3}-p_{k}^{2})$, and $\mathbf{y}_{k}=\mathbf{z}_{k}\times\mathbf{x}_{k}$. 
$O_{k}^{a}$ is the weighted sum of $\left\{ p_{k}^{i}\right\}$, and $\alpha_{k}^{i}$ is the weight coefficient.

\subsubsection{Physical Constraints}
\label{sec_method_physical_constraints}
Given the static equilibrium assumption~\cite{CUP08_ClassicalMechanics}, all forces acting on the object must collectively satisfy Newtonian physical constraints. These constraints are used both to train the force prediction module and to guide the pose aggregation process.

\textbf{Force balance:} the sum of all external forces acting on it must be zero. This leads to the force balance constraint,
\begin{equation}
    \label{equ_force_balance}
    \mathcal{L}_{force}=\left\Vert\sum_{k=1}^{32}F_{k}+\mathbf{G}\right\Vert_{2}^2,
  \end{equation}
where $\mathbf{G}$ represents the gravity vector.
Following~\cite{SigGraph22_Physical}, the magnitude of $\mathbf{G}$ is set to a relative value of 1N.

\textbf{Torque balance:}
the sum of all torques acting on it must be zero. Thus, the torque balance constraint is formulated as,
\begin{equation}
  \label{equ_torque_balance}
  \mathcal{L}_{torque}=\left\Vert\sum_{k=1}^{32}F_{k}\times r_{k}\right\Vert_{2}^2,
\end{equation}
Here $r_{k}$ refers to the position vector of the $k$-th anchor point, 
and $\times$ is a cross product between the two vectors.

\textbf{Contact-force Relation:}
In hand-object interaction, an anchor point can exert force on an object only if it is in contact with the object's surface. Consequently, the magnitude of the contact force should be constrained based on the contact state between the anchor and the object.
The distance between an anchor point and the object's surface serves as a key indicator of this contact state. Specifically, a shorter distance implies a higher likelihood of contact, whereas a larger distance suggests a lower probability of contact. Inspired by this, we approximate the contact-force constraint as
\begin{equation}
  \label{equ_contact_force}
  \mathcal{L}_{contact}=\sum_{k=1}^{32}\Vert F_{k}\Vert_{2}\cdot \left| d_{k} \right|,
\end{equation}
where $\left|d_{k}\right|$ denotes the distance between the anchor point and the object's surface, and $\|F_{k}\|_{2}$ represents the magnitude of the force.
This constraint is an essential physical cue for our physics-based aggregation.

\begin{figure}[t]
    \centering
    \includegraphics[width=0.9\columnwidth]{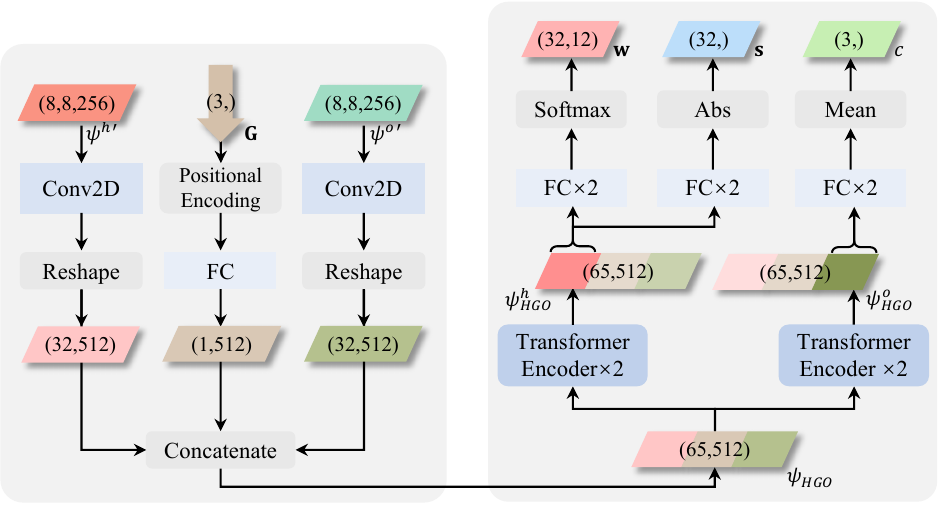}
    \caption {The architecture of our physics network.}
    \label{fig_method_physics_network}
\end{figure}

\subsubsection{Physics Network}

As illustrated in Figure~\ref{fig_method_physics_network}, our physics network takes 1) the hand feature $\psi^{h\prime}$, 2) the object feature $\psi^{o\prime}$, and 3) the gravity vector $\mathbf{G}\in\mathbb{R}^{3}$ as inputs.
During training, $\mathbf{G}$ is aligned vertically downward with respect to the tabletop.
During inference, we approximate $\mathbf{G}$ using the camera's y-axis, as in~\cite{NIPS23_DeepSimHO}.
The network outputs 1) the weight matrix $\mathbf{w}\in\mathbb{R}^{32\times N_{v}}$, 2) the scaling vector $\mathbf{s}\in\mathbb{R}^{32}$, and 3) the object center-of-mass position $c\in\mathbb{R}^{3}$.
The predicted local force $F_k^{\prime}$ is computed using $\mathbf{w}$ and $\mathbf{s}$ as defined in Equation~\ref{equ_local_force}.
The network is supervised using the following loss:
\begin{equation}
  \label{equ_physics_loss}
  \begin{aligned}
    \mathcal{L}_{phy} &= \lambda_{F}\mathcal{L}_{mse}(F^{\prime}, \tilde{F}^{\prime})  +\lambda_{c}\mathcal{L}_{mse}(c, \bar{c}) \\
    &+\lambda_{force}\mathcal{L}_{force}+\lambda_{torque}\mathcal{L}_{torque},
  \end{aligned}
\end{equation}
where $F^{\prime}$ and $\tilde{F}^{\prime}$ are the predicted forces and pseudo force labels, $c$ and $\bar{c}$ are the predicted and ground-truth object center-of-mass positions,
$\lambda_{F}$, $\lambda_{c}$, $\lambda_{force}$, and $\lambda_{torque}$ are hyperparameters balancing the loss terms.
The pseudo force labels $\tilde{F}^{\prime}$ are precomputed via an optimization process constrained by the physical constraints, further details are provided in the extended version.

\begin{figure*}[t]
    \centering
    \includegraphics[width=0.8\textwidth]{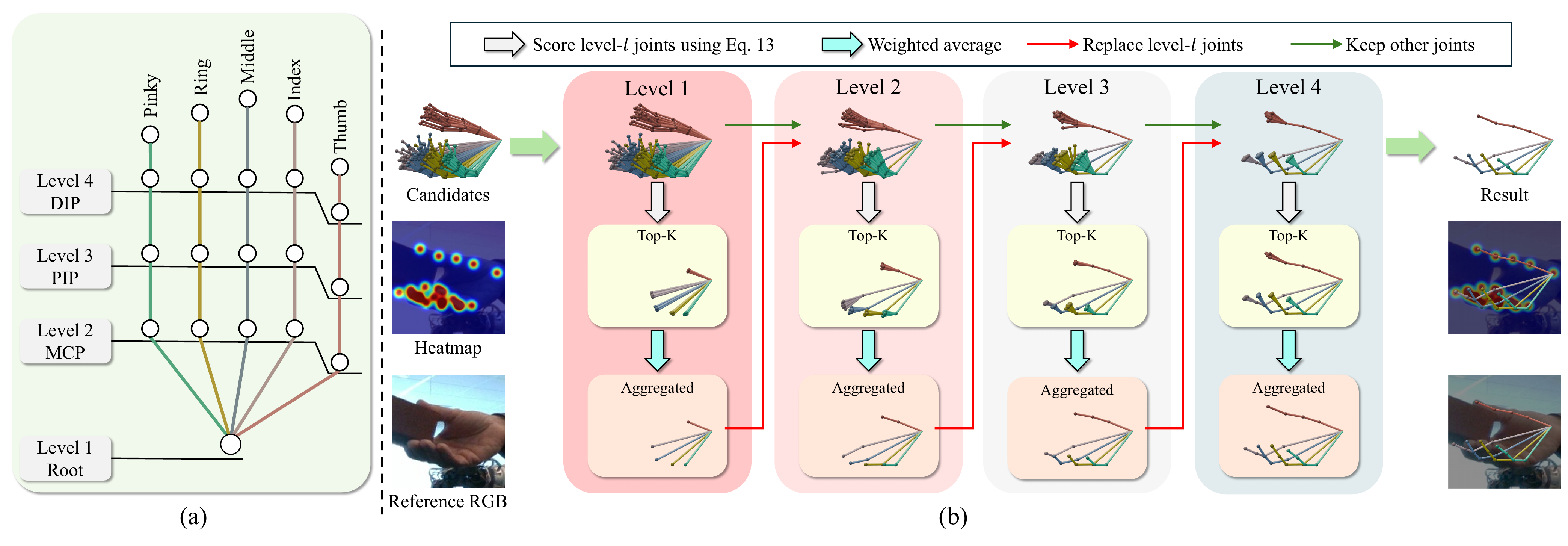} 
    \caption{
        (a) The levels of hand pose parameters;
 	    (b) The visual-based aggregation hierarchically aggregate hand joints from lower to higher levels.}
    \label{fig_method_pose_aggregation}
\end{figure*}

\subsection{Pose Aggregation}
\label{sec_method_pose_aggregation}
Given a set of candidate hand poses $\left\{ (\theta_i, \beta_{reg}) \right\}^N_{i=1}$, where $\theta_i\in\mathbb{R}^{16\times 3}$ are the MANO pose parameters of 16 hand joints and $\beta_{reg}\in\mathbb{R}^{10}$ is the MANO shape parameter, and a set of rigid object pose candidates, where $\left\{ (R_i, T_i) \right\}^N_{i=1}$, $R_i\in\mathbb{R}^{3}$, $T_i\in\mathbb{R}^{3}$ represent rotation and translation respectively, the pose aggregation module is designed to refine these candidates through two sequential stages: visual-based aggregation and physics-based aggregation, as illustrated in Figure~\ref{fig_method_framework}.
The initial pose candidates are generated by a candidate generation module based on a score-based diffusion model~\cite{ICLR21_SDE}. Further details of this module are provided in the extended version.
\subsubsection{Visual-based Aggregation}
\label{sec_method_level_by_level_aggregation}
In articulated systems such as the human hand, higher-level joint positions depend on lower-level ones due to kinematic dependencies. Consequently, errors tend to accumulate across levels, leading to increasingly inaccurate joint predictions at higher levels.
To address this, we propose a level-by-level aggregation strategy that progressively refines joint parameters from lower to higher levels using heatmaps as visual guidance.
As shown in Figure~\ref{fig_method_pose_aggregation}(a), hand joints with degrees of freedom are categorized into four hierarchical levels, with the index set of joints at level $l$ denoted by $L_l$.
The aggregation proceeds iteratively from level 1 to level 4 (Figure~\ref{fig_method_pose_aggregation}(b)). After aggregating each level, the refined joint parameters are used to overwrite the corresponding joints across all candidate poses. This reduces the propagation of errors to higher levels.
For the $j$-th joint in level $L_l$, we compute a visual score $s_i^h$ for each hand pose candidate $(\theta_i, \beta_{reg})$ using the following formula:
\begin{equation}
  \label{equ_hand_visual_score_calculation}
  s_{i}^{h}=\sum_{c\in Children_{j}}H_{c}^{h}\left({\rm Proj2D}^{h}\left(c,\theta_{i},\beta_{reg}\right)\right),
\end{equation}
where ${Children}_j$ denotes the set of higher-level joints within the same kinematic chain as joint $j$, $\text{Proj2D}^h(\cdot)$ denotes the 2D projection of the $c$-th joint for candidate $(\theta_i, \beta_{reg})$, and $H_c^h(\cdot)$ retrieves the heatmap value at the projected 2D location for joint $c$.
Using these scores $\left\{ s_i^h \right\}_{i=1}^N$, the top-K candidates $\left\{ \theta_i \right\}_{i \in K^h}$ are selected. The aggregated pose parameter for joint $j$, denoted by $\tilde{\theta}[j,:]$, is computed as the weighted average of the selected candidates:
\begin{equation}
    \label{equ_hand_visual_score_aggregation}
    \tilde{\theta}[j,:] = \frac{\sum_{i\in K^h} s_i^h \theta_i[j,:]}{\sum_{i\in K^h} s_i^h}.
  \end{equation}
This aggregated value then replaces all $\theta_i[j,:]$ for $i = 1,\dots,N$. This process is repeated for each level until all hand pose parameters are aggregated.

The same principle is applied to object pose aggregation. Object pose parameters are separated into two levels: 1) the translation parameter $T_{o}$, and 2) the rotation parameters $R_{o}$.
The visual score for each object candidate is computed using:
\begin{equation}
    \label{equ_object_visual_score_calculation}
    s_{i}^{o}=\sum_{c=1}^{27}H_{c}^{o}\left({\rm Proj2D}^{o}\left(c,R_{i},T_{i}\right)\right),
  \end{equation}
where $\text{Proj2D}^o(c, R_i, T_i)$ is the 2D projection of the $c$-th object keypoint after applying transformation $(R_i, T_i)$ to the object model, and $H_c^o(\cdot)$ retrieves the heatmap value at the projected 2D location. 
Object translation parameters are aggregated first and used to overwrite the translation parameters of all candidates. Rotation parameters are subsequently aggregated using the updated translations.

\setlength{\tabcolsep}{1mm}
\begin{table}[!t]
	\renewcommand{\arraystretch}{1.1}
	\center
		\begin{tabular}{c|cc|ccc}
			\hline
			\multirow{2}{*}{Method} &
			\multicolumn{2}{c|}{Hand} &
			\multicolumn{3}{c}{Object}
			\cr

			& MJE & PA-MJE & MCE & OCE & ADDS
			\cr
			\hline
			
			Liu et al.~\shortcite{CVPR21_Semi}&
			15.2 & 6.58 & - & - & - 
			\cr
			
			HandOccNet~\shortcite{CVPR22_HandOccNet} &
			14.0 & 5.80 & - & - & - 
			\cr

			H2ONet~\shortcite{CVPR23_H2Onet} &
			14.0 & 5.70 & - & - & - 
			\cr

			HandBooster~\shortcite{CVPR24_HandBooster} &
			11.9 & 5.2 & - & - & - 
			\cr

			SimpleHand~\shortcite{CVPR24_SimpleBaseline} &
			12.4 & 5.5 & - & - & - 
			\cr

			HFL~\shortcite{CVPR23_HFL} &
			12.6 & 5.47 & 48.0 & 42.7 & 33.8 
			\cr

			HOISDF~\shortcite{CVPR24_HOISDF} &
			10.1 & 5.13 & 35.8 & 27.6 & 18.6 
			\cr

      		\hline

            Ours
            & \textbf{10.0} & \textbf{5.08} & \textbf{26.2} & \textbf{23.7} & \textbf{13.5}
			\cr
			
			\hline
			
		\end{tabular}

	\caption{
        Comparison of pose accuracy on \emph{DexYCB Full} (metrics are in mm).
      }
	\label{tab_dexycb_full}
\end{table}

\setlength{\tabcolsep}{3pt}
\begin{table}[!t]
	\renewcommand{\arraystretch}{1.1}
	\center
        \begin{tabular}{c|cc|cc}
            \hline
            \multirow{2}{*}{Method} &
            \multicolumn{2}{c|}{Hand} &
            \multicolumn{2}{c}{Object}
            \cr

			& PA-MJE & MJE & OCE & ADDS
			\cr
			\hline

			Hasson et al.~\shortcite{CVPR19_HOPE} &
			11.0 & - & 67.0 & 22.0 
			\cr

			Hasson et al.~\shortcite{CVPR20_Hasson} &
			11.4 & - & 80.0 & 40.0 
			\cr

			Hampali et al.~\shortcite{CVPR22_KYPT} &
			10.8 & 25.5 & 68.0 & 21.4 
			\cr

			Liu et al.~\shortcite{CVPR21_Semi} &
			10.1 & - & - & - 
			\cr

			DMA~\shortcite{WACV23_DMA} &
			10.1 & 23.8 & 45.5 & 20.8
			\cr

			HFL~\shortcite{CVPR23_HFL} &
			8.9 & 28.9 & 64.3 & 32.4
			\cr

			HandBooster~\shortcite{CVPR24_HandBooster} &
			\textbf{8.5} & 21.1 & - & -
			\cr

			LCP~\shortcite{MM24_LCP} &
			\textbf{8.5} & 21.5 & - & -
			\cr

			HOISDF~\shortcite{CVPR24_HOISDF} &
			9.2 & \textbf{19.0} & 35.5 & 14.4
			\cr

      		\hline

            Ours w/o pretraining &
            8.9 & 21.1 & 29.3 & 15.2
            \cr

			Ours &
			\textbf{8.5} & 19.9 & \textbf{27.1} & \textbf{14.3}
			\cr
			\hline
			
		\end{tabular}
    \caption{
        Comparison of pose accuracy on \emph{HO3Dv2 Full} (metrics are in mm).
        }
	\label{tab_ho3d_full}
\end{table}

\subsubsection{Physics-based Aggregation}
To further enhance the physical plausibility of the hand-object interaction, we introduce a physics-based aggregation step that utilizes physical constraints to guide candidate pose aggregation.
For the hand, we compute a physics-based score as:
\begin{equation}
  \label{equ_hand_physical_score_calculation}
  s^{h}_{phy}=-\mathcal{L}_{force}\cdot\mathcal{L}_{contact}.
\end{equation}
We focus on refining the joints at the highest hierarchy level $L_4$. Let $K_4^h$ denote the top-K hand pose candidates previously aggregated at level $L_4$. We collect joint parameters $\left\{ \theta_i[j,:] \right\}_{i \in K_4^h, j \in L_4}$ and re-rank them based on their physics-based scores computed using Equation~\ref{equ_hand_physical_score_calculation}. The top-K joint parameters $\left\{ \theta_n[j,:] \right\}_{n \in K_{\text{phy}}^h}$ are then averaged to produce the final hand pose estimates for level $L_4$.
For the object, the physics-based score is defined as:
\begin{equation}
  \label{equ_object_physical_score_calculation}
  s^{o}_{phy}=-\mathcal{L}_{torque}\cdot\mathcal{L}_{contact}.
\end{equation}
We retrieve the top-K translation candidates $\left\{ T_i \right\}_{i \in K_T^o}$ and top-K rotation candidates $\left\{ R_j \right\}_{j \in K_R^o}$ from the visual-based aggregation stage. These components are then combined to form a new set of object pose candidates $\left\{ (T_i, R_j) \right\}_{i \in K_T^o, j \in K_R^o}$. Each pair is scored using Equation~\ref{equ_object_physical_score_calculation}, and the top-K combinations $\left\{ (T_n, R_m) \right\}_{(n, m) \in K_{\text{phy}}^o}$ are selected. The final object pose is obtained by averaging these top-ranked pose pairs.
The analysis of the number of candidates and the top-K size is provided in the extended version.

\section{Experiments}
\textcolor{black}{
    Our method is compared against state-of-the-art approaches on two widely used benchmarks: DexYCB~\cite{CVPR21_DexYCB} and HO3D v2~\cite{CVPR20_HO3D}.
}

\textbf{DexYCB}:
To train the model, the training set of official “S0” split~\cite{CVPR21_DexYCB} is employed. 
To evaluate the pose estimation accuracy, two common testing sets are used: 
\emph{DexYCB Full}, the testing set of official “S0” split, includes scenarios of hands approaching objects as well as hands contacting objects; 
To evaluate the physical plausibility of the results, following~\cite{NIPS23_DeepSimHO}, \emph{DexYCB Phy} is employed as the testing set, in which the hands steadily hold the object.

\textbf{HO3Dv2}:
To train the model, the standard training set is employed. 
To evaluate the pose estimation accuracy, standard testing set \emph{HO3Dv2 Full} is used. 
To evaluate the physical plausibility of the results, following~\cite{NIPS23_DeepSimHO,TPAMI24_CPF}, \emph{HO3Dv2 Phy} is employed as the testing set, whose physics plausibility is manually verified by~\cite{TPAMI24_CPF}. 
Some methods use additional data to train their models~\cite{CVPR24_HOISDF,CVPR24_HAMER,CVPR23_H2Onet,WACV23_DMA}. 
Following~\cite{CVPR23_H2Onet}, we optionally pretrain on DexYCB for 5 epochs and report results both with and without this pretraining.

\setlength{\tabcolsep}{0.8mm}
\begin{table}[!t]
	\renewcommand{\arraystretch}{1.1}
	\center
		\begin{tabular}{c|c|c|ccc}
			\hline
			\multirow{2}{*}{Method} &
			\multicolumn{1}{c|}{Hand} &
			\multicolumn{1}{c|}{Object} &
			\multicolumn{3}{c}{Physics}
			\cr

			& MJE$\downarrow$ & SMCE$\downarrow$ & CP(\%)$\uparrow$ & PD$\downarrow$ & SD$\downarrow$
			\cr
			\hline

			Ground Truth &
			- & - & 100 & 9.1 & 6.4
			\cr

			Hasson et al.~\shortcite{CVPR20_Hasson} &
			12.5 & - & 84.35 & 18.0 & 48.3 
			\cr

			DMA~\shortcite{WACV23_DMA} &
			11.5 & - & 89.16 & 15.7 & 35.3 
			\cr

			ArtiBoost~\shortcite{CVPR22_ArtiBoost} &
			10.7 & 16.0 & 94.23 & 15.0 & 27.8 
			\cr

			DeepSimHO~\shortcite{NIPS23_DeepSimHO} &
			11.2 & 17.3 & 95.90 & 14.8 & 24.2 
			\cr

      		\hline

            Ours &
            \textbf{8.5} & \textbf{15.1} & \textbf{98.85} & \textbf{13.4} & \textbf{17.3}
			\cr

			\hline
		\end{tabular}
        \caption{
            Comparison of hand-object physics plausibility on \emph{DexYCB Phy} (metrics, except CP, are in mm).
          }
	\label{tab_dexycb_phy}
\end{table}

\setlength{\tabcolsep}{4pt}
\begin{table}[!t]
	\renewcommand{\arraystretch}{1.1}
	\center
		\begin{tabular}{c|c|ccc}
			\hline

			\multirow{2}{*}{Method} &
			\multicolumn{1}{c|}{Object} &
			\multicolumn{3}{c}{Physics}
			\cr

			& SMCE$\downarrow$ & CP(\%)$\uparrow$ & PD$\downarrow$ & SD$\downarrow$
			\cr
			\hline

			Hasson et al.~\shortcite{CVPR20_HO3D} &
			5.35 & 78.52 & 2.02 & 6.40 
			\cr

			CPF~\shortcite{TPAMI24_CPF} &
			5.74 & 96.47 & 1.65 & 3.16 
			\cr

			DMA~\shortcite{WACV23_DMA} &
			4.79 & 93.07 & 1.88 & 3.47 
			\cr

			ArtiBoost~\shortcite{CVPR22_ArtiBoost} &
			4.86 & 94.47 & 1.27 & 2.83 
			\cr

			DeepSimHO~\shortcite{NIPS23_DeepSimHO} &
			5.28 & 96.64 & 1.17 & 2.42
			\cr

      		\hline

            Ours &
            \textbf{3.12} & \textbf{98.80} & \textbf{0.96} & \textbf{2.21}
			\cr

			\hline
		\end{tabular}
        \caption{
            Comparison of hand-object physics plausibility on \emph{HO3Dv2 Phy} (metrics, except CP, are in cm).
          }
	\label{tab_ho3d_phy}
\end{table}

\begin{table*}[!t]
	\renewcommand{\arraystretch}{1.1}
	\center
		\begin{tabular}{ccc|cc|cc|ccc}
			\hline

			Force &
			Visual-based &
			Physics-based &
			\multicolumn{2}{c|}{Hand} &
			\multicolumn{2}{c|}{Object} &
			\multicolumn{3}{c}{Physics}
			\cr

			Prediction & Aggregation & Aggregation & MJE$\downarrow$ & PA-MJE$\downarrow$ & OCE$\downarrow$ & ADDS$\downarrow$ & CP(\%)$\uparrow$ & PD$\downarrow$ & SD$\downarrow$
			\cr
			\hline

			 & & &
            12.54 & 5.45 & 35.30 & 20.14 & 95.42 & 14.4 & 30.4 
            \cr

			\checkmark & & &
			12.16 & 5.33 & 29.81 & 16.66 & 96.51 & 14.1 & 25.9 
			\cr

			\checkmark & \checkmark & &
			10.05 & 5.09 & 26.12 & 15.21 & 97.95 & 13.9 & 20.0
			\cr

			\checkmark & \checkmark & \checkmark &
			\textbf{10.01} & \textbf{5.08} & \textbf{23.72} & \textbf{13.47} & \textbf{98.85} & \textbf{13.4} & \textbf{17.3}
			\cr

			\hline
		\end{tabular}

	\caption{
    	Ablation study (metrics, except CP, are in mm). Hand and object metrics are evaluated on \emph{DexYCB Full}. Physics metrics are evaluated on \emph{DexYCB Phy}.
  	}
	\label{tab_ablation}
\end{table*}

\begin{figure}[!t]
	\centering
	\includegraphics[width=0.9\columnwidth]{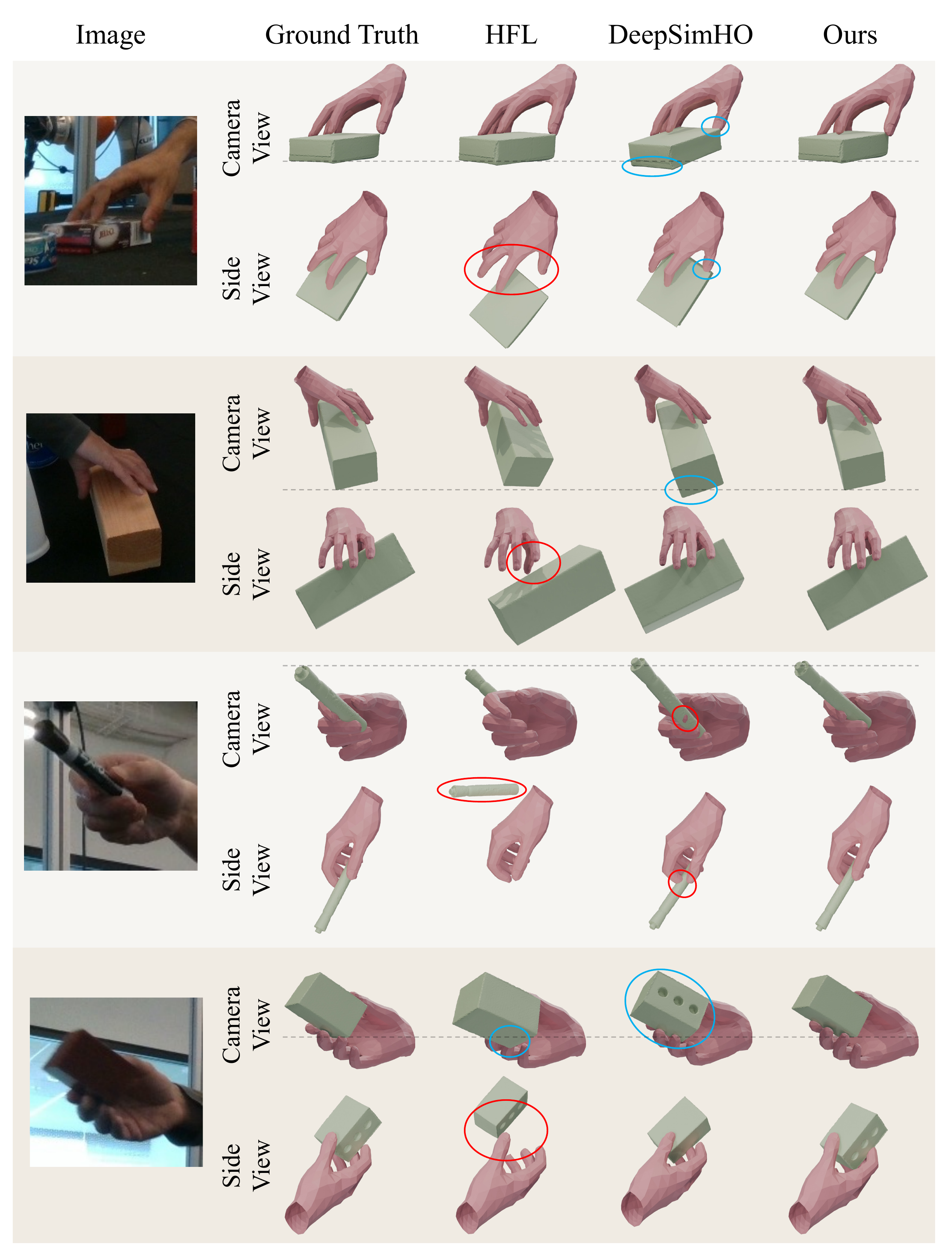}
	\caption{
		Qualitative results on \emph{DexYCB Phy}. Red circles indicate incorrect hand object interaction. Blue circles point out the incorrect pose estimation. For more qualitative results, please refer to the extended version.
	}
	\label{fig_qualitative}
\end{figure}

\subsection{Evaluation Metrics}

\textbf{Pose Metrics:} For hand, Mean Joint Error (MJE) and Procrustes aligned Mean Joint Error (PA-MJE) are reported. 
For object, Object Center Error (OCE), Mean Corner Error (MCE), Symmetry-aware Mean Corner Error (SMCE), and average closest point distance (ADD-S) are evaluated.

\noindent\textbf{Physics Metrics:} The contact percentage (CP)is calculated to assess the ratio of predictions with hand-object contact. 
The penetration depth (PD) is used to measure the maximum penetration distance between hand and object predictions. 
To evaluate the stability of hand holding object, the simulation displacement (SD)~\cite{NIPS23_DeepSimHO} is used to compute the average object center displacement after 200ms in the virtual physical simulator.

\subsection{Pose Estimation Accuracy}
The proposed method is compared with the state-of-the-art pose estimation methods, including both hand-object pose estimators~\cite{CVPR19_HOPE,MM24_LCP,3DV21_Hasson,CVPR23_gSDF,CVPR22_ArtiBoost,WACV23_DMA,CVPR20_Hasson,CVPR21_Semi,CVPR23_HFL,CVPR24_HOISDF} and hand pose estimators~\cite{CVPR22_HandOccNet,CVPR22_KYPT,CVPR24_MS_MANO,CVPR23_H2Onet,CVPR24_HandBooster,CVPR24_SimpleBaseline}.

The results on \emph{DexYCB Full} are shown in Table~\ref{tab_dexycb_full}. The proposed method outperforms the compared methods on both hand and object estimation. 
Especially in terms of object metrics, the proposed method significantly outperforms the second best with reducing the error of MCE, OCE and ADDS by 26.8\%, 14.1\%, and 27.4\%, respectively.  

The results on \emph{HO3Dv2 Full} are shown in Table~\ref{tab_ho3d_full}. 
Compared with the state-of-the-art methods, the proposed method achieves better performance on joint hand-object pose estimation.
For the hand-related metrics, the proposed method is comparable to state-of-the-art methods. For object-related metrics, the proposed method achieves the best performance.

\subsection{Physical Plausibility}
\label{sec_physical_plausibility}
The proposed method is compared with physics-based methods~\cite{NIPS23_DeepSimHO,TPAMI24_CPF,3DV21_Hasson}, and visual-based methods~\cite{CVPR22_ArtiBoost,WACV23_DMA}.

The results on \emph{DexYCB Phy} are shown in Table~\ref{tab_dexycb_phy}. Among the existing methods, DeepSimHO~\cite{NIPS23_DeepSimHO} achieves the best physical performance, but its accuracy on hand and object is suboptimal. 
Comparing to DeepSimHO, ArtiBoost~\cite{CVPR22_ArtiBoost} has better pose estimation accuracy, but its physical plausibility is inferior. The existing methods cannot balance on both physics and pose estimation. In contrast, the proposed method significantly outperforms existing methods in both physical plausibility and pose accuracy.

The results on \emph{HO3Dv2 Phy} are shown in Table~\ref{tab_ho3d_phy}. 
Our method achieves state-of-the-art performance across all physics metrics. Compared to the strongest baseline, DeepSimHO, our method further reduces PD and SD while improving contact rate and object accuracy, demonstrating the effectiveness of our approach in producing both physically plausible and accurate poses.

The qualitative results are shown in Figure~\ref{fig_qualitative}. The visual-based method HFL~\cite{CVPR23_HFL} appears to have good visual consistency in the camera view. 
However, in the side view, incorrect hand-object interactions are observed. 
Compared to HFL, the physics-based method DeepSimHO~\cite{NIPS23_DeepSimHO} has more plausible hand-object interaction but less visual consistency in the front view. 
Comparing to these methods, the proposed method achieves better performance on both visual consistency and physical plausibility. 
For more qualitative results, please refer to the extended version.

\subsection{Ablation Study}
\label{sec_ablation_study}

The experimental results of ablation study are shown in Table~\ref{tab_ablation}. 
Starting from the baseline with no force prediction or aggregation modules, the model exhibits the weakest performance across hand, object, and physics metrics. 
Introducing the force prediction module alone yields modest improvements, particularly in object pose accuracy (OCE drops from 35.30 mm to 29.81 mm, and ADD-S from 20.14 mm to 16.66 mm), indicating that learning physical cues already aids hand-object pose quality.       
Adding the visual-based aggregation further improves both hand and object metrics (MJE reduced to 10.05 mm, OCE to 26.12 mm), demonstrating its effectiveness in reducing error accumulation.
Finally, the inclusion of the physics-based aggregation module yields the best overall results. It achieves the lowest errors across all metrics, particularly in physics-based evaluations: CP increases from 97.95\% to 98.85\%, while PD and SD drop from 13.9 mm to 13.4 mm and from 20.0 mm to 17.3 mm, respectively. These gains confirm the effectiveness of the proposed physics-based aggregation strategy in enhancing physical plausibility for hand-object pose estimation results.
Additional analyses of aggregation strategies, \textcolor{black}{inference-time efficiency}, and the influence of candidate quantity and Top-K selection are provided in the extended version.

\section{Conclusion}
In this paper, we have presented a novel approach for hand-object pose estimation that effectively combines visual and physical cues to tackle the inherent challenges of visual consistency and physical plausibility in existing methods.
Our method learns not only 2D visual representations but also 3D physical cues, using a semi-supervised learning strategy combined with a local-to-global transformation mechanism.
The predicted visual and physical cues guide a candidate pose aggregation module, enabling the aggregation of physically plausible and visually coherent hand-object poses.
Our experimental results demonstrate that the proposed framework achieves state-of-the-art performance in both pose estimation accuracy and physical plausibility.
\textcolor{black}{
    Future work includes incorporating temporal information to model dynamic equilibrium in hand-object interactions and leveraging physical cues to improve object reconstruction.
}

\newpage

\section{Acknowledgments}
\textcolor{black}{
    This work was supported by the National Natural Science Foundation of China under Grant 62273318,
    the Science and Technology Project funds of Power Construction Corporation of China Ltd., 
    and the Science and Technology Project of Sinohydro Bureau 8 Company Ltd.
}

\bibliography{aaai2026}

\twocolumn[%
\begin{center}
\textbf{\LARGE Supplementary Material}
\end{center}
\vspace{0.5cm}
]

\begin{abstract}
In this supplementary material, we provide additional details about our proposed method, including visualizations of the pseudo force labels and an overview of the diffusion models used to generate hand and object pose candidates. We also describe the hyperparameter settings and analyze the impact of the number of candidates and the top-K selection size on performance. Furthermore, we present extended experiments, such as comparisons of different aggregation strategies and an in-depth analysis of our proposed aggregation scheme. We additionally report comprehensive quantitative metrics for both hand and object pose estimation.
Finally, we provide a detailed breakdown of the inference runtime of each module in our framework.
\end{abstract}


\section{Pseudo Force Label}
\subsection{Pseudo Force Label Generation}
To generate the pseudo force labels, we employ physical constraints to optimize the weight matrix $\textbf{w}\in\mathbb{R}^{32\times N_v}$ and the scaling vector $\textbf{s}\in\mathbb{R}^{32}$.
To enforce the conditions $\sum_{j=1}^{N_v} w_{ij} = 1$ and $s_k \geq 0$, we optimize instead the matrix $\tilde{\textbf{w}}\in\mathbb{R}^{32\times N_v}$ and the vector $\tilde{\textbf{s}}\in\mathbb{R}^{32}$. The relation between these variables are expressed as follows:
\begin{equation}
  {\bf w_k}={\rm softmax}({\bf \tilde{w}_k}), \quad s_k = \vert \tilde{s}_k \vert,
\end{equation}
where ${\bf w}_k$ and ${\bf \tilde{w}_k}$ denote the $k$-th column of $\textbf{w}$ and $\tilde{\textbf{w}}$, respectively, while $s_k$ and $\tilde{s}_k$ refer to the $k$-th element of $\textbf{s}$ and $\tilde{\textbf{s}}$, respectively.

\begin{figure}[!t]
  \centering
  \includegraphics[width=3.2in]{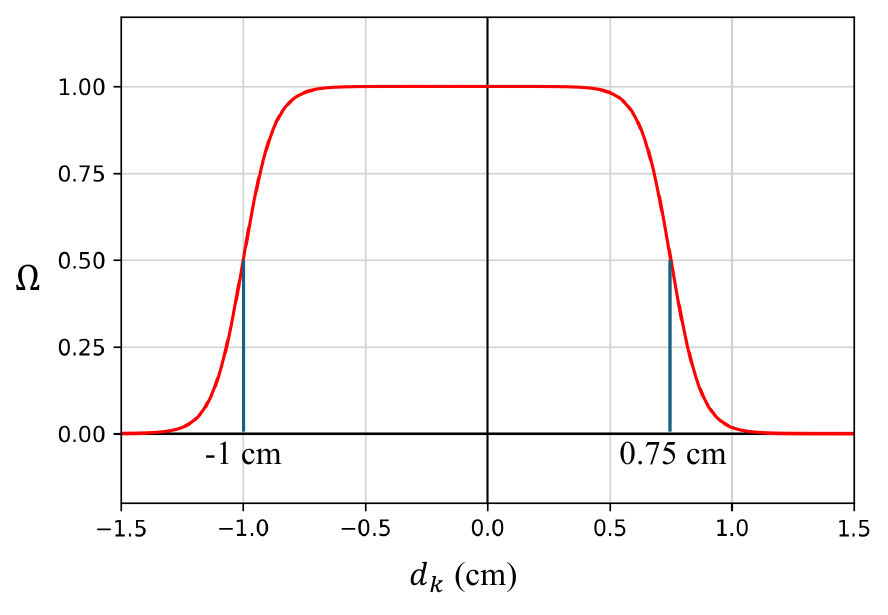}
  \caption{
    The mapping function $\Omega(d_k)$.
  }
  \label{fig_mapping_function}
\end{figure}

The initial values for $\tilde{\textbf{s}}$ are determined based on the distance between the hand and the object. 
Specifically, given the hand and object meshes, we compute the signed distance $d_k$ from the object's surface to the anchor point $k$. 
A mapping function $\Omega$ is defined to smoothly converts the signed distance $d_k$ to a value between $(0, 1)$:
\begin{equation}
  \Omega(d_k) = \frac{1}{(1+e^{-16(d_k+1)})(1+e^{-16(d_k-0.75)})}.
\end{equation}
This function transitions smoothly to 1 for $d_k\in(0, 0.75)$ and approaches 0 otherwise, as illustrated in Fig.~\ref{fig_mapping_function}.
If $\Omega(d_k) < 0.1$, we set $\tilde{s}_k$ to a fixed value $0$ and keep it unchanged; otherwise, we initialize $\tilde{s}_k$ to a value of 0.05.
The initial values for $\{\tilde{w}_{k,j}\}$ are set uniformly to $\frac{1}{N_v}$.

The optimization process includes two phases.
In the first phase, the primary objective is to optimize the matrix $\tilde{\textbf{w}}$, ensuring that the resultant hand force vector acts against the object's gravity. For this purpose, the physical constraint $\mathcal{L}_{force}$ is employed:
\begin{equation}
  \mathcal{L}_{force} = \left\Vert\sum_{k=1}^{32} \left(R_k^{L2G} s_k \sum_{j=1}^{N_v} w_{k,j} \mathbf{v}_j\right) - \mathbf{G}\right\Vert_2^2.
\end{equation}
The optimization goal for this phase can be formulated as:
\begin{equation}
  \mathop{arg\min}\limits_{\tilde{\mathbf{w}}} \mathcal{L}_{force}.
\end{equation}

In the second phase, the goal is to minimize the resultant force and torque exerted on the object.
Additionally, the process encourages anchor points that are in contact with the object to exert force while reducing the force exerted by anchor points that are farther away.
To achieve this goal, the matrix $\tilde{\textbf{w}}$ and the vector $\tilde{\textbf{s}}$ are jointly optimized using the loss functions $\mathcal{L}_{force}$, $\mathcal{L}_{torque}$, and $\mathcal{L}_{contact2}$:
\begin{equation}
  \mathop{arg\min}\limits_{\tilde{\mathbf{w}}, \tilde{\mathbf{s}}} \left( \mathcal{L}_{force} + 30\mathcal{L}_{torque} + 0.1\mathcal{L}_{contact2} \right).
\end{equation}
Here, $\mathcal{L}_{force}$ is used to minimize the resultant force. $\mathcal{L}_{torque}$ is used to minimize the resultant torque which is defined as:
\begin{equation}
  \mathcal{L}_{torque} = \left\Vert\sum_{k=1}^{32} \left(R_k^{L2G} s_k \sum_{j=1}^{N_v} w_{k,j} \mathbf{v}_j\right) \times {r}_k\right\Vert_2^2.
\end{equation}
The function $\mathcal{L}_{contact2}$ is used to encourage anchor points in contact with the object to exert force while suppressing the force from those far away. It is defined as:
\begin{equation}
  \mathcal{L}_{contact2} = \sum_{k=1}^{32} \log^2\left(
    \frac{\Omega(d_k) \sqrt{\sum_{i=1}^{32} s_i^2}}{s_k \sqrt{\sum_{i=1}^{32} \Omega^2(d_i)} + \epsilon}
    \right),
\end{equation}
where $\epsilon$ is a hyperparameter, set to $10^{-5}$, to avoid division by zero.
In our experiments, we utilize the AdamW optimizer with a learning rate of $10^{-3}$.
Typically, convergence for the first phase occurs after approximately 300 optimization steps, while the second phase converges after about 2700 steps.

\subsection{Visualization of Pseudo Force and Predicted Force}
\begin{figure}[!t]
	\centering
	\includegraphics[width=1\columnwidth]{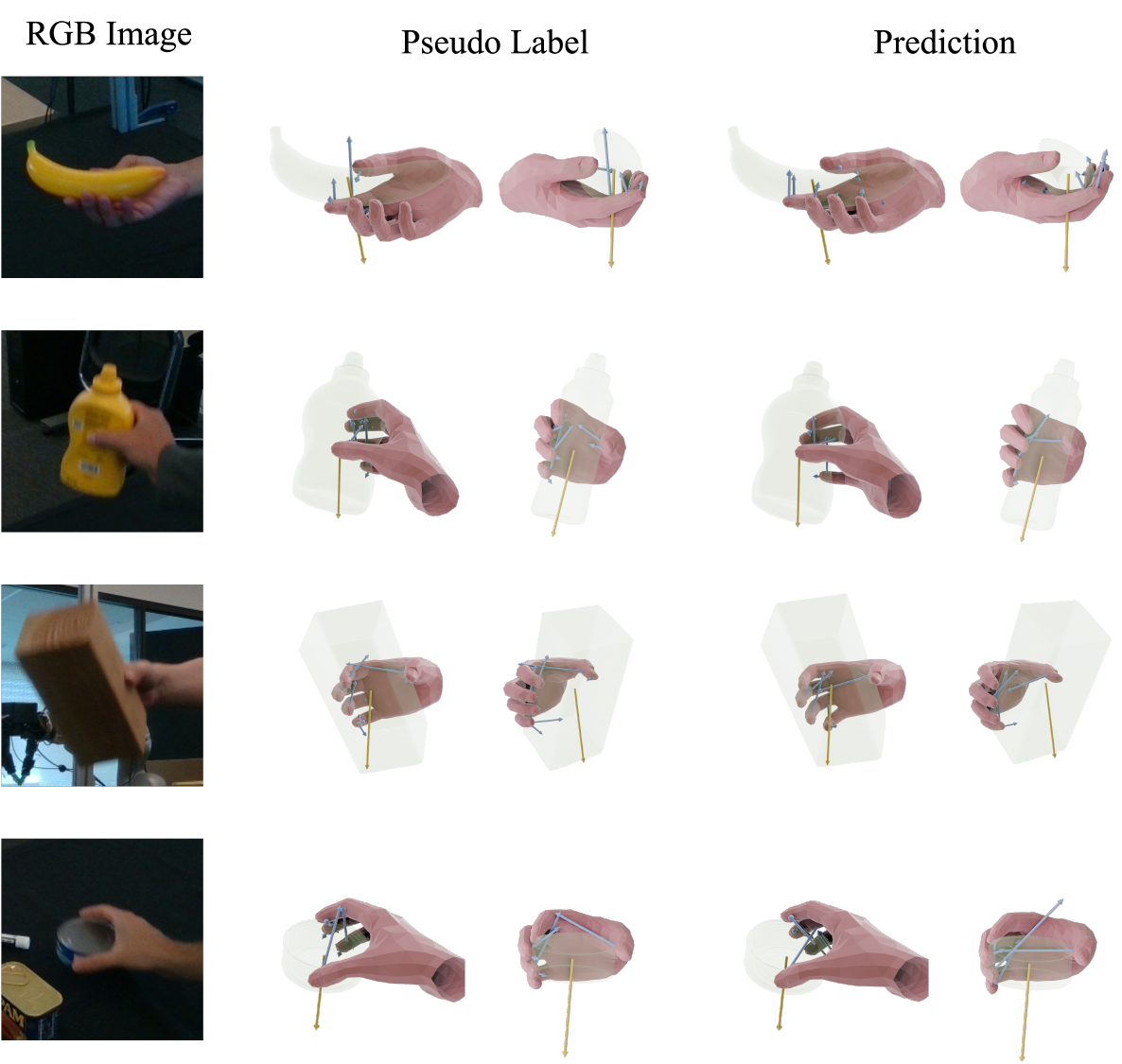}
	\caption{
	  The visualization of the pseudo force label and predicted force. Blue arrows indicate the force vectors, while yellow arrows indicate the gravity vectors.
	}
	\label{fig_pseudo_force_visualization}
  
  \end{figure}

The visualization of the pseudo force and predicted force is shown in Figure~\ref{fig_pseudo_force_visualization}.

\begin{figure}[!t]
	\centering
	\includegraphics[width=2.2in]{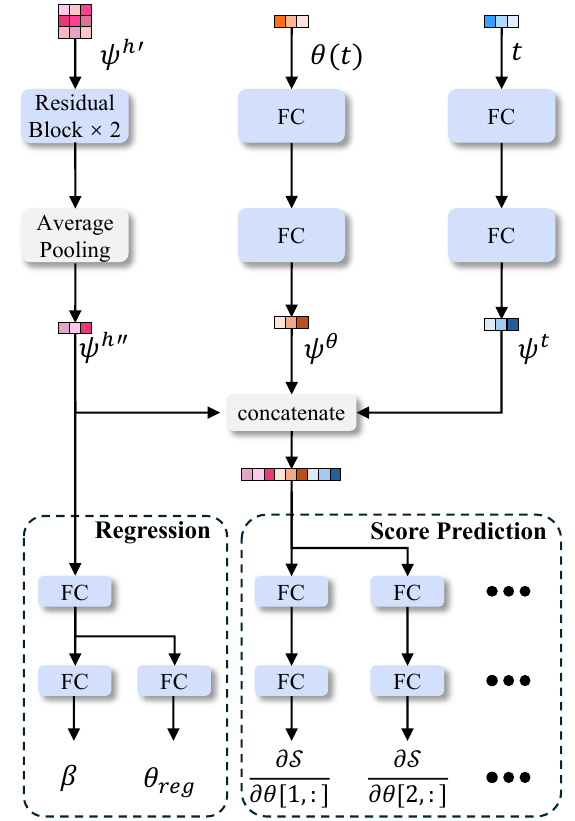}
	\caption{
	  The network architecture of our hand candidate generator.
	}
	\label{fig_hand_candidate_generator}
  
  \end{figure}

  \setlength{\tabcolsep}{3pt}
  \begin{table*}[!t]
      \renewcommand{\arraystretch}{1.3}
      \center
          \begin{tabular}{c|cc|cc|cc|cc|cc}
              \hline
              Num. of &
              \multicolumn{2}{c|}{Top-K Size in VA} &
              \multicolumn{2}{c|}{Top-K Size in PA} &
              \multicolumn{2}{c|}{Hand} &
              \multicolumn{2}{c|}{Object} &
              \multicolumn{2}{c}{Complexity}
              \cr
  
              Candidates & Hand & Object & Hand & Object & MJE & PA-MJE & OCE & ADDS & GFLOPs & Params (M)
              \cr
              \hline
  
              50 & 30 & 10 & 5 & 5 &
              10.12 & 5.14 & 24.71 & 14.03 & 22.41 & 48.95 
              \cr

              100 & 30 & 10 & 5 & 5 &
              10.01 & 5.08 & 23.72 & 13.47 & 25.51 & 48.95 
              \cr

              200 & 30 & 10 & 5 & 5 &
              10.00 & 5.07 & 22.75 & 12.89 & 31.94 & 48.95
              \cr

              \hline
              
              100 & 10 & 10 & 5 & 5 &
              10.23 & 5.24 & 23.74 & 13.48 & 25.51 & 48.95 
              \cr

              100 & 30 & 30 & 5 & 5 &
              10.01 & 5.08 & 24.66 & 13.68 & 25.51 & 48.95 
              \cr

              100 & 60 & 60 & 5 & 5 &
              10.05 & 5.10 & 26.29 & 14.33 & 25.51 & 48.95 
              \cr

              \hline

              100 & 30 & 10 & 1 & 1 &
              10.04 & 5.11 & 24.17 & 13.50 & 25.51 & 48.95 
              \cr

              100 & 30 & 10 & 5 & 5 &
              10.01 & 5.08 & 23.72 & 13.47 & 25.51 & 48.95 
              \cr

              100 & 30 & 10 & 10 & 10 &
              10.05 & 5.08 & 24.40 & 14.34 & 25.51 & 48.95 
              \cr
  
              \hline
          \end{tabular}
      \caption{
          Analysis of candidate number and top-K size on \emph{DexYCB Full}.
        }
      \label{tab_analysis_hyper_parameters}
  \end{table*}

\section{Candidate Generation}

To generate hand candidates, a candidate generator $\Theta^{h}$ produces a set of plausible MANO hand poses $\{\theta_{i}\}_{i=1}^{N}$, where each $\theta_{i}\in\mathbb{R}^{16\times6}$ represents the continuous 6D rotation representation of the joint angles for the wrist and 15 finger joints.
Following previous efforts~\cite{CVPR22_KYPT, CVPR23_HFL, CVPR24_HOIST}, the wrist position is assumed to be known.
%

The candidate generators for both hands and objects are modeled based on the score-based diffusion model~\cite{ICLR21_SDE, NIPS19_SDE}.
In what follows, we focus on describing the hand candidate generator, as the object candidate generator operates in a similar manner.
The diffusion process $\{\theta(t)\}_{t=0}^{1}$ is constructed using the Variance-Exploding Stochastic Differential Equation (VE-SDE)~\cite{ICLR21_SDE},
\begin{equation}
  \label{equ_ve_sde}
  {\rm d}\theta(t)=\sqrt{\frac{{\rm d}\left(\sigma^{2}(t)\right)}{{\rm d}t}{\rm d}w}.
\end{equation}
Here, $w$ is a Brownian motion, $t$ denotes the time variable, $\sigma(t)=\sigma_{\min}(\frac{\sigma_{\max}}{\sigma_{\min}})^t$, $\sigma_{\min}$ and $\sigma_{\max}$ are hyperparameters set to 0.01 and 50, respectively. During training, the score network $\Theta^{h}\left(\theta(t),t|\psi^{h\prime}\right)$ learns the score equation $\nabla_{\theta}\log p_{t}\left(\theta(t)|\psi^{h\prime}\right)$, where $p_{t}\left(\theta(t)|\psi^{h\prime}\right)$ is the marginal distribution of $\theta(t)$ at time $t$. Based on Denoising Score Matching (DSM)~\cite{NC11_SocreMatching}, the loss function used to train the score network is 
\begin{equation}
  \label{equ_diff_loss}
  \mathcal{L}_{diff}^{h}=\sigma^{2}(t)\cdot \left\Vert\Theta^{h}\left(\theta(t),t|\psi^{h\prime}\right)-\frac{\theta(0)-\theta(t)}{\sigma^{2}(t)}\right\Vert_{2}^{2},
\end{equation}
where $t\sim\mathcal{U}(\epsilon,1)$ with the hyperparameter $\epsilon\rightarrow0$ set to $10^{-5}$. During inference, we sample $N$ pose candidates $\{\theta_{i}(t_{f})\}_{i}^{N}$ from a Gaussian distribution $\mathcal{N}\left(\mathbf{0},\sigma^{2}(t_{f})\mathbf{I}\right)$. For each sample $\theta_{i}(t_{f})$, the final candidate $\theta_{i}(0)$ is obtained by solving the probability flow ODE (PF-ODE)~\cite{ICLR21_SDE} from $t=t_{f}$ to $t=\epsilon$:
\begin{equation}
  \label{equ_pf_ode}
  \frac{{\rm d}\theta}{{\rm d}t}=-\sigma(t)\dot{\sigma}(t)\nabla_{\theta}\log p_{t}(\theta|\psi^{h\prime}),
\end{equation}
where $t_{f}$ is a hyperparameter set to 0.55 for human hands and 0.65 for objects in our experiments. The PF-ODE is solved using the RK45 ODE solver~\cite{JCAM80_RK45ODE}.

As illustrated in Figure~\ref{fig_hand_candidate_generator}, the hand candidate generator network $\Theta^h$ takes as input 1) the hand feature $\psi^{h \prime}$, 2) the sampled pose $\theta(t)\in\mathbb{R}^{16\times6}$, and 3) the time variable $t$.
The processing steps are as follows: 1) $\psi^{h\prime}$ is passed through two residual blocks, then reshaped into a feature vector $\psi^{h\prime\prime}$.
2) $\theta(t)$ is processed through two fully connected layers, yielding the feature vector $\psi^{\theta}\in\mathbb{R}^{256}$.
3) $t$ is encoded into the a feature vector $\psi^{t}\in\mathbb{R}^{128}$~\cite{ICLR21_SDE}.
Then these three feature vectors, $\psi^{h\prime\prime}$, $\psi^t$, and $\psi^\theta$ are concatenated and passed through multiple fully connected layers to predict the score of each joint $\{\frac{\partial S}{\partial \theta[j,:]}\}_{j=1}^{J^h}$, with $S=\log p_t(\theta(t)|\psi^h)$.
In addition to skeletal 3D hand pose $\theta$, the MANO hand model also contains the hand shape parameter vector $\beta\in\mathbb{R}^{10}$.
It is predicted by a regression branch, consisting of three fully connected layers, that takes $\psi^{h\prime\prime}$ as input and outputs the hand shape parameter $\beta$ as well as the pose parameter $\theta_{reg}$.

For object candidate generation, an object candidate generator $\Theta^{o}$ produces a set of plausible object poses $\{\phi_{i}\}_{i=1}^{N}$. Here $\phi_{i}\in\mathbb{R}^{9}$ contains the object pose parameters $(R,T)$, with $R\in\mathbb{R}^{6}$ representing the continuous 6D rotation parameters, and $T\in\mathbb{R}^{3}$ the translation parameters.
The object candidate generator network follows a similar architecture but does not include a regression branch. Instead, it outputs gradient scores $\{\frac{\partial S}{\partial R}, \frac{\partial S}{\partial T}\}$ in the score prediction branch.

\section{Implementation Details}
\subsection{Hyper-parameters Settings}
The number of hand joints $|J^h|$ is set to 21, following the MANO model.
The number of object keypoints $|K^o|$ is set to 27, consisting of one center point, eight corner points, twelve edge midpoints, and six face midpoints.
The heatmap dimensions $h_m$ and $w_m$ are set to 64.
The friction coefficient $\mu$ is set to 1.0.
The number of base vectors $N_v$ is set to 12.
The loss weights are configured as follows: $\lambda_{hmap} = 10^{3}$, $\lambda_{F} = 10$, $\lambda_{c} = 10^2$, $\lambda_{force} = 1$, $\lambda_{torque} = 30$, $\lambda_{\theta} = 1$, $\lambda_{\beta} = 10$, and $\lambda_{V} = 10^{4}$.
The candidate generator module produces 100 pose candidates for both the hand and the object.
In visual-based aggregation, the top-K size is set to 30 for the hand and 5 for the object. In physics-based aggregation, the top-K size is 10 for both the hand and the object.

\subsection{Training Details}
The input images are cropped to (256, 256) pixels, and the batch size is set to 64. 
The data augmentation strategies include random rotation, translation, color jetting and Gaussian noise. 
The network parameters are optimized with the AdamW optimizer. 
The learning rate is initially set to 1e-4 and decays exponentially with an exponent of 0.98. 
The overall loss is formulated as $\mathcal{L}=\mathcal{L}_{hmap}+\mathcal{L}_{Phy}+\mathcal{L}_{pose}$. 
Typically, the overall loss converges after 45 epochs. 
During both training and testing phases, we determine to apply physical constraints or not, based on whether the hand is grasping the object and lifting it off the table.
If the object is lifted more than 5~cm above the tabletop, apply the physical constraints for hand-object interaction;
conversely, if the distance remains within 5~cm off the tabletop, we assume it is primarily supported by the table, thus do not apply the physical constraints.
The experiments are conducted on a PC with two RTX 4090 GPUs. 

\subsection{Influence of Candidate Number and Top-K Size}

We investigate the impact of the number of pose candidates and the Top-K selection sizes in both visual-based aggregation (VA) and physics-based aggregation (PA) on overall performance and computational complexity, as shown in Table~\ref{tab_analysis_hyper_parameters}. Increasing the number of candidates from 50 to 200 consistently improves both hand and object accuracy, with the mean joint error (MJE) decreasing from 10.12 to 10.00 and the object ADD-S improving from 14.03 to 12.89, albeit at the cost of increased GFLOPs. Notably, expanding the Top-K size in VA from 10 to 60 while keeping PA fixed degrades object performance (ADDS rises from 13.48 to 14.33), suggesting that overly large visual candidate pools may introduce noise. In contrast, adjusting the Top-K size in PA shows a marginal effect on performance, indicating that PA is relatively robust to this hyperparameter. 
Based on this analysis, a balanced configuration using 100 candidates, VA Top-K sizes of 30 for the hand and 10 for the object, and PA Top-K sizes of 5 for both hand and object achieves strong performance across both hand and object metrics, without incurring additional computational cost.

\setlength{\tabcolsep}{2pt}
\begin{table}[!t]
	\renewcommand{\arraystretch}{1.3}
	\center
		\begin{tabular}{c|ccc|ccc}
			\hline

			\multirow{2}{*}{Method} &
			\multicolumn{3}{c|}{Hand} &
			\multicolumn{3}{c}{Object}
			\cr
			\cline{2-7}

			& MJE & PA-MJE & MME & MCE & OCE & ADDS
			\cr
			\hline

			Random &
			17.11 & 6.87 & 16.47 & 45.82 & 41.50 & 23.78 
			\cr

			Baseline 1 &
			12.26 & 5.32 & 11.90 & 35.88 & 32.57 & 28.37 
			\cr

			Baseline 2 &
			12.27 & 5.26 & 11.01 & 34.71 & 31.62 & 17.80 
			\cr

			Baseline 3 &
			12.69 & 5.56 & 12.21 & 30.01 & 25.53 & 14.94 
			\cr

      		\hline

            VA &
            10.05 & 5.09 & 9.77 & 28.63 & 26.12 & 15.21
			\cr

			VA + PA &
			\textbf{10.01} & \textbf{5.08} & \textbf{9.73} & \textbf{26.23} & \textbf{23.72} & \textbf{13.47}
			\cr

			\hline
		\end{tabular}
	\caption{
    	Analysis of aggregation strategy on \emph{DexYCB Full} (metrics, except CP, are in mm).
  	}
	\label{tab_analysis_aggregation}

\end{table}

\section{Analysis of Aggregation Strategy}

The effectiveness of the proposed candidates aggregation strategy is compared with the following baselines: 
\emph{Random} denotes to randomly select a candidate. 
\emph{Baseline 1} denotes to average all candidates. 
\emph{Baseline 2} denotes the highest likelihood solution. 
\emph{Baseline 3} denotes sorting based on the distance between the predicted 2D joints and the projected 2D joints of the candidates, and then averaging the top-K candidates with smaller average distances. 
\emph{VA} denotes the visual cue-based aggregation strategy. 
\emph{VA + PA} denotes the proposed aggregation strategy. 
The number of candidates is set to 100. The size of top-K is set to 30 for hand and 10 for object.
MME denotes the mean mesh error.

Experimental results are shown in Table~\ref{tab_analysis_aggregation}. 
\emph{Random} directly uses a random candidate, which includes a lot of noises, leading the worst performance. 
\emph{Baseline 1} and \emph{Baseline 2} use simple aggregation strategies, significantly improving the hand-object pose estimation accuracy.
It suggests that candidate aggregation is important for a multi-hypotheses method. 
\textit{Baseline 3} attempts to incorporate visual cues by leveraging predicted 2D joint locations for candidate selection. However, it performs worse than \textit{Baseline 1} and \textit{Baseline 2} for hand estimation. This degradation is primarily due to frequent occlusions during hand-object interactions, which impair the reliability of 2D joint predictions.
In contrast, our \textit{visual-based} aggregation strategy leverages heatmaps as visual cues to progressively refine hand pose selection. This approach leads to substantial improvements in hand pose accuracy while maintaining comparable object estimation performance to \textit{Baseline 3}.
Finally, our full method (\textit{Ours}) combines visual-based aggregation and physics-based aggregation. This strategy achieves the best overall performance across both hand and object metrics, demonstrating the complementary strengths of visual and physical reasoning in candidate aggregation.

\begin{table}[t]
    \centering
    \renewcommand{\arraystretch}{1.2}
    \setlength{\tabcolsep}{4pt}
    \begin{tabular}{lc|ccc}
    \hline
    Metric & Joint Index & Right & Left & Both \\
    \hline
    MJE  & 1  & 3.89  & 4.11  & 4.00  \\
    MJE  & 2  & 6.71  & 6.92  & 6.82  \\
    MJE  & 3  & 8.88  & 9.14  & 9.01  \\
    MJE  & 4  & 12.28 & 12.73 & 12.50 \\
    MJE  & 5  & 7.51  & 7.67  & 7.59  \\
    MJE  & 6  & 9.33  & 9.78  & 9.55  \\
    MJE  & 7  & 10.72 & 11.20 & 10.96 \\
    MJE  & 8  & 13.16 & 14.08 & 13.61 \\
    MJE  & 9  & 7.72  & 8.04  & 7.88  \\
    MJE  & 10 & 10.01 & 10.28 & 10.15 \\
    MJE  & 11 & 11.72 & 11.79 & 11.76 \\
    MJE  & 12 & 15.12 & 14.95 & 15.04 \\
    MJE  & 13 & 7.37  & 7.74  & 7.55  \\
    MJE  & 14 & 9.77  & 9.92  & 9.84  \\
    MJE  & 15 & 12.14 & 11.88 & 12.01 \\
    MJE  & 16 & 16.26 & 15.48 & 15.88 \\
    MJE  & 17 & 7.62  & 8.09  & 7.85  \\
    MJE  & 18 & 9.73  & 10.26 & 9.99  \\
    MJE  & 19 & 11.96 & 12.31 & 12.14 \\
    MJE  & 20 & 16.11 & 15.99 & 16.05 \\
    MJE     & all & 9.91  & 10.11 & 10.01 \\
    PA-MJE & all & 5.16  & 4.99  & 5.08  \\
    \hline
    \end{tabular}
    \caption{Detailed hand pose accuracy on \emph{DexYCB Full} dataset.}
    \label{tab:mje_metrics}
    \end{table}
    
    \begin{table*}[t]
        \centering
        \renewcommand{\arraystretch}{1.2}
        \setlength{\tabcolsep}{3pt}
        \begin{tabular}{l|rrrrrrr}
            \hline
            Object & MCE & OCE & ADD & ADDS & ADD$_{0.1d}$ & ADDS$_{0.1d}$ & REP \\
            \hline
            002\_master\_chef\_can   & 20.74 & 19.74 & 43.06 & 10.31 & 34.12\% & 91.83\% & 21.44 \\
            003\_cracker\_box        & 23.99 & 22.64 & 25.04 & 12.00 & 72.18\% & 94.84\% & 8.02  \\
            004\_sugar\_box          & 24.54 & 21.81 & 28.76 & 12.12 & 59.03\% & 89.15\% & 10.18 \\
            005\_tomato\_soup\_can   & 22.14 & 21.02 & 32.23 & 11.09 & 23.73\% & 80.85\% & 13.90 \\
            006\_mustard\_bottle     & 18.75 & 17.31 & 22.64 & 9.11  & 70.39\% & 94.44\% & 9.30  \\
            007\_tuna\_fish\_can     & 21.66 & 20.98 & 35.59 & 11.16 & 15.82\% & 77.40\% & 17.03 \\
            008\_pudding\_box        & 37.98 & 36.28 & 43.38 & 25.44 & 52.82\% & 85.95\% & 183.34 \\
            009\_gelatin\_box        & 24.78 & 23.39 & 28.30 & 12.57 & 38.29\% & 78.09\% & 9.52  \\
            010\_potted\_meat\_can   & 21.60 & 20.31 & 30.15 & 10.96 & 40.26\% & 83.32\% & 13.65 \\
            011\_banana              & 28.61 & 23.04 & 34.98 & 13.75 & 48.11\% & 84.99\% & 14.08 \\
            019\_pitcher\_base       & 35.09 & 33.32 & 41.05 & 18.29 & 57.81\% & 91.82\% & 16.77 \\
            021\_bleach\_cleanser    & 26.70 & 22.72 & 32.74 & 12.65 & 64.30\% & 92.56\% & 14.67 \\
            024\_bowl               & 21.00 & 19.80 & 65.25 & 11.22 & 22.24\% & 92.32\% & 36.71 \\
            025\_mug                & 20.24 & 19.86 & 29.91 & 9.41  & 37.47\% & 91.21\% & 13.01 \\
            035\_power\_drill       & 29.88 & 28.18 & 32.34 & 14.73 & 58.98\% & 88.99\% & 11.57 \\
            036\_wood\_block        & 29.76 & 24.86 & 95.41 & 14.34 & 7.20\%  & 88.93\% & 55.23 \\
            037\_scissors           & 34.19 & 28.43 & 38.84 & 17.82 & 34.48\% & 75.57\% & 14.56 \\
            040\_large\_marker      & 25.39 & 22.20 & 29.06 & 14.65 & 10.52\% & 67.41\% & 11.33 \\
            052\_extra\_large\_clamp & 33.74 & 26.21 & 56.29 & 15.64 & 29.09\% & 87.04\% & 27.86 \\
            061\_foam\_brick        & 23.32 & 21.67 & 44.41 & 11.73 & 11.31\% & 72.20\% & 22.06 \\
            \hline
            average\_instance       & 26.23 & 23.72 & 39.20 & 13.47 & 39.65\% & 85.46\% & 26.40 \\
            \hline
            \end{tabular}
        \caption{Per-object performance on \emph{DexYCB Full} dataset.}
        \label{tab:object}
        \end{table*}

\section{Detailed Performance on \emph{DexYCB Full} Dataset}
We provide a detailed performance analysis of the hand pose accuracy on the \emph{DexYCB Full} dataset, including separate evaluations for the left hand, right hand, and both hands, as well as per-joint accuracy. The results are summarized in Table~\ref{tab:mje_metrics}.
Additionally, Table~\ref{tab:object} reports per-object performance on the \emph{DexYCB Full} dataset.
For object pose estimation, we also evaluate the following metrics: Average Distance of Model Points (ADD, mm); ADD with a threshold of 10\% of the object's diameter (ADD${0.1d}$); average distance to the closest model point with a 10\% diameter threshold (ADDS${0.1d}$); and Reprojection Error (REP, pixel).

\subsection{Analysis of Inference Time}

\begin{table}[t]
    \centering
    \renewcommand{\arraystretch}{1.2}
    \setlength{\tabcolsep}{3pt}
    \begin{tabular}{l c}
    \hline
    \textbf{Module} & \textbf{Time (ms)} \\ 
    \hline
    Feature extraction & 18.6 \\
    Candidate Generation (hand) & 34.5 \\
    Candidate Generation (object) & 25.6 \\
    Visual-based aggregation (hand) & 18.9 \\
    Visual-based aggregation (object) & 6.75 \\
    Physics-based aggregation (hand) & 10.6 \\
    Physics-based aggregation (object) & 4.1 \\
    \hline
    \textbf{Total} & \textbf{119.05} \\
    \hline
    \end{tabular}
    \caption{Runtime analysis of each module in our framework.}
    \label{tab:inference_time}
\end{table}

As shown in Table~\ref{tab:inference_time}, the main computational bottleneck lies in the Candidate Generation stage, which relies on a score-based diffusion model to produce pose candidates. We acknowledge that diffusion-based models tend to have higher inference costs compared to deterministic regressors.
However, our key contribution lies in the visual-physical representation learning and visual–physical aggregation framework, both of which are agnostic to the choice of candidate generator. This modular design allows our approach to seamlessly integrate with more efficient generators (e.g., VAE or flow-based models), which can substantially reduce inference time without altering the overall framework. We believe this flexibility enhances the practical scalability and general applicability of our method.

\end{document}